%% file: main.tex
\documentclass[10pt]{IEEEtran}
\usepackage{amsmath,amsfonts}
\usepackage{etoolbox}
\usepackage{array}
\usepackage{pifont}
\usepackage[caption=false,font=normalsize,labelfont=sf,textfont=sf]{subfig}
\usepackage{textcomp}
\usepackage{stfloats,multicol}
\usepackage{url}
\usepackage{verbatim}
\usepackage{graphicx}
\usepackage{cite}
\usepackage{float}
\usepackage{adjustbox}
\usepackage{caption}
\usepackage[breaklinks,colorlinks,colorlinks=true]{hyperref}
\usepackage[capitalize]{cleveref}

\usepackage{booktabs,makecell, multirow, tabularx, tabularray}
\usepackage{latexsym}
\usepackage{algorithmic,algorithm}
\usepackage{subcaption}
\usepackage{graphbox}
\usepackage{xspace} 

\usepackage{xcolor}

\usepackage{lipsum}

\makeatletter
\DeclareRobustCommand\onedot{\futurelet\@let@token\@onedot}
\def\@onedot{\ifx\@let@token.\else.\null\fi\xspace}
\def\eg{\emph{e.g}\onedot}

\def\etal{\emph{et al}\onedot}

\makeatother
\begin{document}

\title{PolyFootNet: Extracting Polygonal Building Footprints in Off-Nadir Remote Sensing Images}

\author{
        Kai Li, \IEEEmembership{Student Member, IEEE},
        Yupeng Deng,
        Jingbo Chen,~\IEEEmembership{Member, IEEE},
        Yu Meng*,\thanks{* Yu Meng is the corresponding author. This research was funded by the National Key R\&D Program of China under Grant number 2021YFB3900504.} 
        \\
        Zhihao Xi,
        Junxian Ma,\thanks{Kai Li, Junxian Ma, and Chenhao Wang are with Aerospace Information Research Institute, Chinese Academy of Sciences, Beijing 100101, China, and also with School of Electronic, Electrical and Communication Engineering, University of Chinese Academy of Sciences, Beijing 100049, China.}
        \thanks{Kai Li, Maolin Wang, and Xiangyu Zhao are also with Applied Machine Learning Lab, School of Data Science, City University of Hong Kong, Kowloon Tong, Hong Kong 999077.}
        Chenhao Wang,\thanks{Yupeng Deng, Jingbo Chen, Yu Meng, and Zhihao Xi are with Aerospace Information Research Institute, Chinese Academy of Sciences, Beijing 100101, China}
        Maolin Wang,
        Xiangyu Zhao
}

\markboth{Journal of \LaTeX\ Template, Under Review paper, October~2024}%
{Shell \MakeLowercase{\textit{et al.}}: A Sample Article Using IEEEtran.cls for IEEE Journals}


\maketitle
\input{sec/0_abstract}    
\input{sec/1_intro.tex}
\input{sec/X_rela}
\input{sec/2_method}
\input{sec/3_exp}
\input{sec/4_ablation}
\input{sec/5_discussions}

\input{sec/6_conclusion}
\bibliographystyle{IEEEtran}
\bibliography{main} 

\input{author/authors}

\end{document}

%% file: sec/0_abstract.tex
\begin{abstract}
\input{chapters/Abstract}
\end{abstract}


%% file: chapters/Abstract.tex
Extracting polygonal building footprints from off-nadir imagery is crucial for diverse applications. 
Current deep-learning-based extraction approaches predominantly rely on semantic segmentation paradigms and post-processing algorithms, limiting their boundary precision and applicability. 
However, existing polygonal extraction methodologies are inherently designed for near-nadir imagery and fail under the geometric complexities introduced by off-nadir viewing angles. 
To address these challenges, this paper introduces Polygonal Footprint Network (PolyFootNet), a novel deep-learning framework that directly outputs polygonal building footprints without requiring external post-processing steps. PolyFootNet employs a High-Quality Mask Prompter to generate precise roof masks, which guide polygonal vertex extraction in a unified model pipeline. 
A key contribution of PolyFootNet is introducing the Self Offset Attention mechanism, grounded in Nadaraya-Watson regression, to effectively mitigate the accuracy discrepancy observed between low-rise and high-rise buildings. 
This approach allows low-rise building predictions to leverage angular corrections learned from high-rise building offsets, significantly enhancing overall extraction accuracy. 
Additionally, motivated by the inherent ambiguity of building footprint extraction tasks, we systematically investigate alternative extraction paradigms and demonstrate that a combined approach of building masks and offsets achieves superior polygonal footprint results. 
Extensive experiments validate PolyFootNet's effectiveness, illustrating its promising potential as a robust, generalizable, and precise polygonal building footprint extraction method from challenging off-nadir imagery. 
To facilitate further research, we will release pre-trained weights of our offset prediction module at \url{https://github.com/likaiucas/PolyFootNet}.

%% file: chapters/Keywords.tex
Building footprint extraction, Building detection, Segment Anything Model (SAM), Off-nadir aerial image, Nadaraya-Watson regression, Oblique monocular images

%% file: sec/1_intro.tex
\input{chapters/Introduction}

%% file: chapters/Introduction.tex
\section{Introduction}
\label{sec:Intro}
\IEEEPARstart{B}{uilding} Footprint Extraction (BFE) in off-nadir images has been a research subject for over a decade, forming the foundation for critical tasks such as 3D building reconstruction and building change detection~\cite{bfe_review, obli_1, obli_2, obli_3}. Off-nadir imagery is particularly attractive as it provides a more efficient and economical alternative to near-nadir images, enabling broader coverage with fewer acquisitions.
Early approaches to BFE primarily consisted of two categories: geometric-feature-based algorithms and traditional machine learning algorithms~\cite{structureapproch, geofeature, feature_bfe_1,feature_bfe_2, svmbfe}. 
Then, the rise of deep convolutional networks revolutionized BFE, leading to various deep learning approaches. Among these innovations, offset-based methods have emerged as a prominent solution in recent years, demonstrating remarkable performance in building feature extraction~\cite{cnnbfe2020, MTBRNet, pang2023detecting, BONAI, MLSBRN, obm, WSMTBRNet,EnhanceBFETGRS}. 

\begin{figure}
	\centering
	\includegraphics[width=0.9\linewidth]{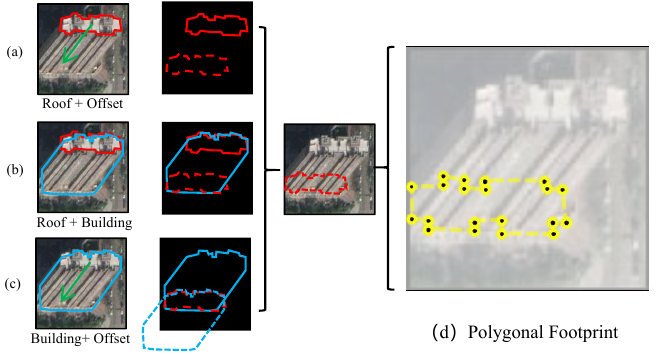}
	\caption{Previous approaches primarily relied on Paradigm (a) for extracting building footprints. In this study, we explore the extraction of building footprints using task decomposition based on Paradigms (b) and (c). Additionally, this work achieves the first implementation of Paradigm (d), enabling the direct extraction of polygonal footprints by the model via decoding \textit{offset tokens} and \textit{vertex tokens}. Compared to Paradigms (a), (b), and (c), which depend on post-processing algorithms to extract final results, the proposed method eliminates this dependency. 
    }
	\label{fig.intro}
\end{figure}

However, former building footprint methods mainly focus on semantic segmentation following~Fig.\ref{fig.intro}~(a), which is often restricted in accurately delineating boundaries and exhibits limited generalizability, which can affect
their real-world applicability. 
Meanwhile, to the best of our knowledge, existing polygonal BFE methods are designed for near-nadir scenarios~\cite{sampolybuild, hisup}, relying on a crucial hypothesis that the projected polygonal positions of the visible roof and invisible footprint significantly overlap—a condition that no longer holds true in off-nadir imagery.

Results of the latest methods~\cite{MTBRNet, BONAI, MLSBRN} followed~Fig.\ref{fig.intro}~(a), which can automatically extract footprints, failed to be applied as polygonal footprints because instance segmentation methods struggled to identify a single building footprint for each building since their processing methods, Region Proposal Network (RPN) and Non-Maximum Suppression (NMS) will assign more than one mask result in pieces for one building~\cite{obm}. 
These repeatedly predicted footprints are hard to post-process with other image features since the footprints on off-nadir images are less visible, especially the tall buildings, compared with well-overlapped footprints with roofs in near-nadir images. 
Meanwhile, the performance gap between predicting bungalows and tall buildings was discovered, especially in the prediction of roof-to-footprint offsets. This made the predicted polygonal footprints of bungalows inaccurately express their footprints.



Therefore, beyond merely establishing a polygonal footprint extraction paradigm tailored for off-nadir imagery, two critical issues warrant further investigation: (1) mitigating the performance discrepancy observed between predictions for low-rise (bungalows) and high-rise buildings and (2) systematically exploring multiple plausible solutions inherent to the BFE problem, as illustrated by examples in Fig.\ref{fig.intro}(b) and (c).

In this paper, we proposed PolyFootNet, which can extract polygonal building footprints from off-nadir images as Fig.\ref{fig.intro}~(d). 
To the best of our knowledge, PolyFootNet is the first paradigm that achieves polygonal building footprints without post-processing, such as OpenCV\footnote{https://opencv.org} operations, which were commonly used in LOFT~\cite{BONAI} and Offset Building Model (OBM)~\cite{obm}.

\begin{figure*}
  \centering
  \includegraphics[width=\linewidth]{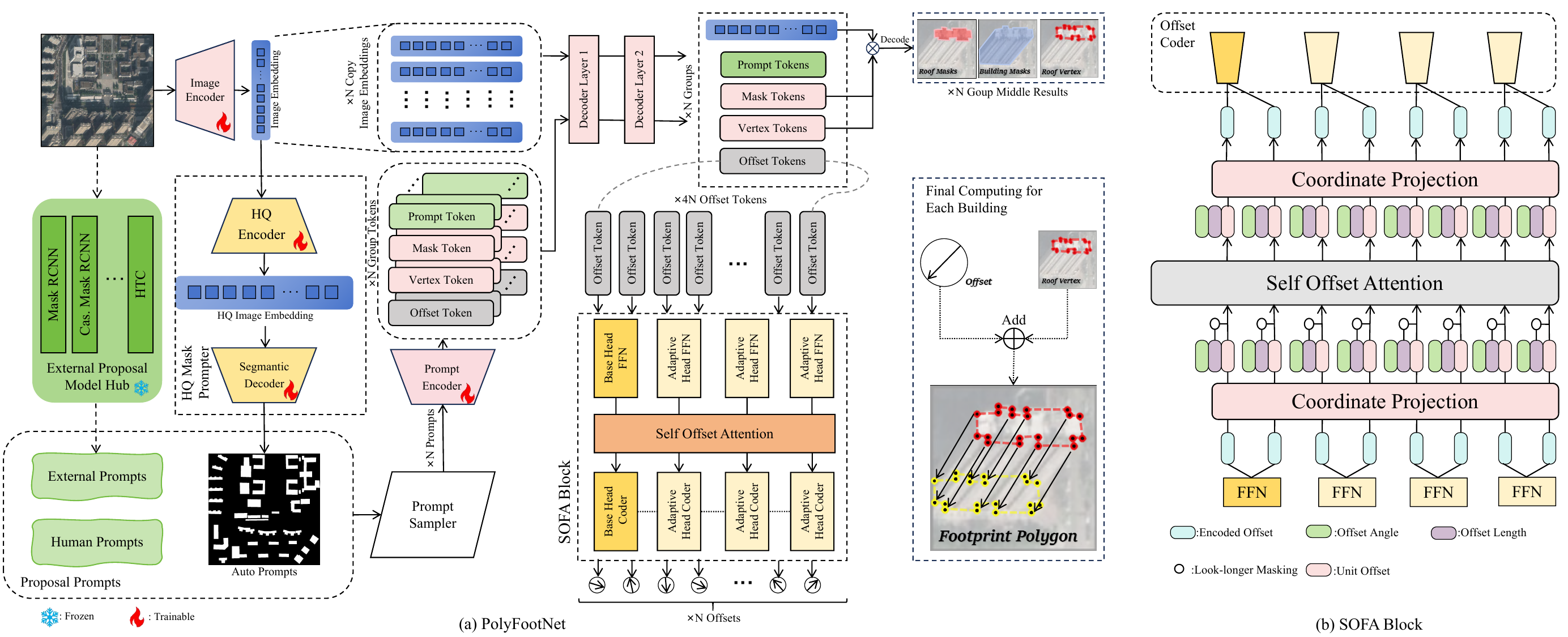}
  \caption{
    This figure illustrates the main structures of PolyFootNet and SOFA. 
    In (a), PolyFootNet's newly added Proposal Networks allow the model to extract buildings automatically.
    In the prompt level, a roof vertex task was added, and the model can directly compute the location of the roof vertex. 
    The footprint polygon is calculated directly on the coordinate. 
    In (b), we provide a detailed SOFA Block. 
    Once outputted from the Feed Forward Network (FFN), the encoded offset will be fed to SOFA. 
    Then, adjusted offsets will be passed to offset coders and compute the final output offsets. 
    }
  \label{fig.main}
\end{figure*}

Specifically, as shown in Fig.\ref{fig.main}~(a), PolyFootNet extracts polygonal building footprints by directly computing the footprint vertices in the coordinates. 
PolyFootNet is a model built on the Segment Anything Model (SAM)~\cite{sam}, which supports zero-shot segmentation. 
To avoid misrepresentation caused by repetitive predictions from the RPN, PolyFootNet employs a High-Quality Mask Prompter to generate roof masks as automatic prompts, subsequently enabling its decoder to extract and compute the polygonal footprints. Additionally, PolyFootNet is flexible enough to leverage pre-trained models from existing libraries for roof segmentation prompts or directly utilize human visual prompts to provide roof-to-footprint offsets for each roof, thus facilitating downstream applications and future research. 

To bridge the performance gap among different buildings within the image, we specifically designed a Self Offset Attention (SOFA) mechanism, formulated using Nadaraya-Watson Regression~\cite{nadaraya, watson, chinpangho_kr_2}, based on earlier observations~\cite{obm} that low buildings exhibit accurate length predictions but poor angular predictions, whereas high-rise buildings demonstrate accurate angular predictions. 
From the experimental results, the proposed Self-Offset Attention (SOFA) mechanism enables low-rise buildings to perform self-correction by utilizing angle offsets learned from high-rise building predictions, significantly improving their overall polygon extraction accuracy.

On the other hand, motivated by the inherent ambiguity and multi-solution nature of extracting building footprints, we explored alternative solutions, as illustrated in Fig.\ref{fig.intro}. Our experiments revealed that adopting a prediction scheme based on the combination of \textit{building masks + offsets} effectively improves the accuracy of polygonal footprint extraction.

In summary, the contributions of this paper are as follows:
\begin{enumerate}
    \item[$\bullet$] We proposed PolyFootNet, the first polygonal footprint extraction network, demonstrating strong generalization capability.
    
    \item[$\bullet$] To bridge the performance gap between different buildings, we designed SOFA based on the Nadaraya-Watson regression. This module can improve the angle quality of low-rise building offsets. 

    \item[$\bullet$] We explore the multi-solution of BFE and discover that \textit{building mask + offset} maybe more suitable for extract building footprint masks.

    \item[$\bullet$] The experiments on three datasets, BONAI, OmniCity-View3 and Huizhou, demonstrate the practical effectiveness of PolyFootNet.
\end{enumerate}

%% file: sec/X_rela.tex

\section{Related Work}
\input{chapters/Rela}

%% file: chapters/Rela.tex
PolyFootNet is a multi-task network built upon the architecture of the SAM~\cite{sam}, integrating tasks such as prompt segmentation, polygon extraction, and offset learning. 
The SOFA block is designed using the Nadaraya-Watson regression framework and an attention mechanism. 
This paper also explores the multiplicity of solutions for BFE by leveraging existing geometric relationships, geographical knowledge, and iterative methods to enhance performance.

In the following subsections, we will review related works in the application of SAM, polygonal building extraction, and building offset methods to highlight the novelty and contributions of PolyFootNet.

\subsection{Segment Anything Model and its application}
The SAM~\cite{sam} is a foundational model for segmentation, supporting tasks using point, bounding box, and semantic prompts. 
SAM~2~\cite{sam2}  was introduced for promptable image and video segmentation, offering faster inference speeds and improved accuracy on both image and video tasks compared to the original SAM. 
Point to Prompt (P2P)~\cite{P2P}, also based on SAM, transforms point supervision into fine visual prompts through a two-stage iterative refinement process. 
Additionally, GaussianVTON~\cite{chen2024gaussianvton} employed a SAM-based model for post-editing view images after face refinement. 
SAM-HQ~\cite{samhq} enhanced the quality of SAM’s image embeddings by incorporating deconvolution blocks, while OBM~\cite{obm} introduced offset tokens and the ROAM structure to enable SAM to extract footprint masks.

In this paper, PolyFootNet builds upon the idea of SAM-HQ to generate high-quality semantic prompts for automatic extraction. 
Furthermore, PolyFootNet introduces a novel concept of shallow vertex tokens to extract key points for footprints. These vertex tokens are related to roof polygons.

\subsection{Polygonal mapping of buildings}
Polygonal mapping of buildings involves extracting vectorized building instances that accurately represent building edges. 
Douglas~\etal~\cite{douglas1973algorithms} introduced the Douglas-Peucker simplification techniques, but these algorithms often produce rough results that fail to capture the high-quality edges of buildings. 
Wei~\etal~\cite{wei2019toward} proposed refinement strategies based on empirical building shapes, while Girard~\etal~\cite{FrameField} used Frame-Field methods to better align extracted fields with ground truth contours. Zorzi~\etal~\cite{zorzi2021machine} described all building polygons in an image as an undirected graph, connecting detected vertices to form the building boundaries. Hisup~\cite{hisup} addressed the challenge of mask reversibility by using deep convolutional neural networks for vertex extraction, followed by boundary tracing of predicted building segmentations to connect the vertices. SAMPolyBuild~\cite{sampolybuild} is a model for solving building layout results in near lidar scenarios, which achieves vectorized result extraction through cropping of pre-selected boxes and editing of weights.

In this paper, PolyFootNet introduces the concept of vertex tokens for extracting vertices and these vertices. 
Different from SAMPloyBuild, PolyFootNet will not crop each building to ensure each token can reference global information~\cite{obm}. Meanwhile, to address the consequential imbalance between positive and negative samples, we propose the Dynamic Scope Binary Cross Entropy Loss (DS-BCE Loss).
Unlike the aforementioned methods, PolyFootNet is the first model to extract polygonal building footprints from off-nadir images. 
The footprint polygons are derived through vector operations, offering higher precision.

\subsection{Offset-based footprint extraction}
Extracting building footprints through offset-based methods leverages the structural similarity between roofs and footprints. 
Christie \etal~\cite{christie2020unet} proposed a method using a U-Net decoder to predict image-level orientation and pixel-level height values. 
Building upon this, Li \etal~\cite{MTBRNet, MLSBRN,WSMTBRNet}
introduced multi-task learning approaches for the BFE problem, enabling models to train on datasets with varying labels. 
LOFT~\cite{BONAI} was later developed to extract building footprints as part of an instance segmentation task, representing a building footprint with a roof mask and roof-to-footprint offset. 
OBM~\cite{obm} was the first model to tokenize offset representations and extract building footprints using SAM.
However, these models all only extract semantic mask results and require a sequence of post-processing operations to derive footprint masks. 

In this paper, we expand on traditional offset-based methods, solving the BFE problem by integrating multiple tasks rather than relying solely on offsets. 
Prior knowledge is applied to explore the potential of using various sources of information. 
For instance, we investigate how building footprints can be extracted through building segmentation and offsets or building and roof segmentation. 
Finally, PolyFootNet is integrated with other models to enable fully automatic building footprint extraction.

\subsection{Offset building model}
The OBM~\cite{obm} is a model inherited from the SAM~\cite{sam}. 
The primary contribution of OBM is introducing the concept of Offset Tokens and designing associated encoding and decoding methods, which have technically brought the issues related to Off-Nadir imagery into the transformer era. This model supports interactive footprint extraction with visual prompts. 
OBM proposed a Reference Offset Augmentation Module (ROAM), a module including a base head and adaptive heads, following the idea of Mix-of-Expert~\cite{jacobs1991moe}.
The encoding and decoding methods are included in Base Head and Adaptive Head, consisting of one Feed Forward Network and an Offset Coder. By fine-tuning SAM, OBM achieves precise roof segmentation in interactive mode. Furthermore, integrating the orientation awareness brought by Offset Tokens enables the direct segmentation of building footprints.

Another contribution in this paper is that the authors discovered from the experimental results that longer offsets have better directions than shorter offsets. 
PolyFootNet designed Self Offset Attention based on this discovery to mitigate the angle performance of shorter offsets. 

%% file: sec/2_method.tex
\input{chapters/Methodology}

%% file: chapters/Methodology.tex
\section{Methodology}
\label{sec:Method}
This section outlines the methodology of this study. 
We begin by reintroducing the BFE problem, and then introduce the core designs of PolyFootNet.

\subsection{Problem statement}
\label{sec:1}
In an off-nadir remote sensing image \textit{I}, there are $N$ buildings represented as $\hat{B}={\hat{b_1}, \hat{b_2}, \dots, \hat{b_N}}$. 
The BFE problem identifies the building footprints $F={f_1, f_2, \dots, f_N}$. 
In OBM, each building $\hat{b_i}$ is represented by a corresponding prompt $p_i$, which is interactively fed into the model along with the image \textit{I}. 
The model then predicts the roof segmentation $r_i$ and the roof-to-footprint offset $o_i$. The footprint $f_i$ is derived by applying the offset to the roof mask in the evaluation stage.

In PolyFootNet, the process is enhanced with the capability for fully automatic footprint prediction, allowing $p_i$ to be empty. 
PolyFootNet introduces a roof vertex segmentation task to extract vertex points $v_i$ for each building $\hat{b_i}$, and re-focuses on building body segmentation $b_i$. 
Additionally, in PolyFootNet, prompts are primarily designed for roof-related tasks, differing from OBM by reducing semantic overlap in off-nadir scenarios.

In summary, PolyFootNet introduces two additional prompt-level segmentation tasks along with a global semantic segmentation task for prompting:
(1) prompt-level roof vertex segmentation task; (2) prompt-level building segmentation; (3) roof semantic segmentation. 
\begin{figure}
\centering
\includegraphics[width=1.0\linewidth]{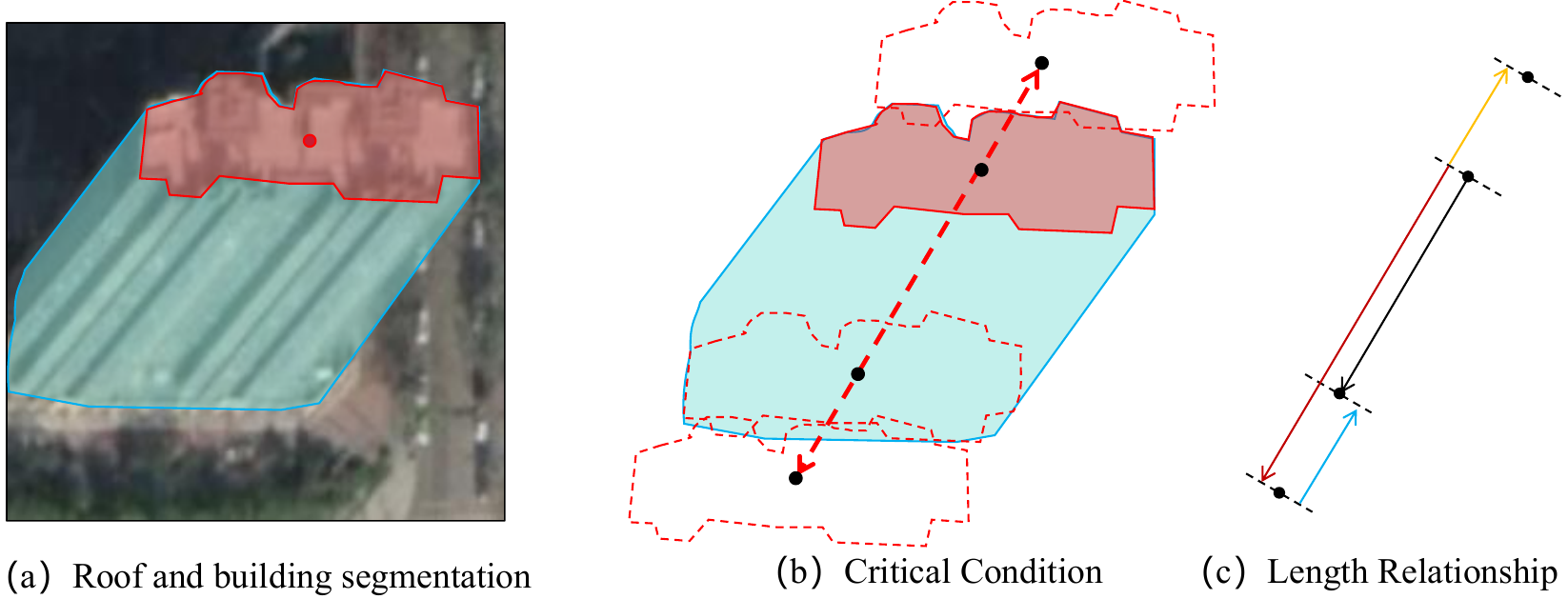}
\caption{(a) describes the predicted roof and building for one building. (b) displays the critical condition of regressing building offset (c) abstracts the situation of (b).}
\label{fig.edge}
\end{figure}

\subsection{Overview of PolyFootNet}
\label{sec:3}
{{Fig.\ref{fig.main}}~(a)} illustrates the architecture of the proposed Polygonal Footprint Network (PolyFootNet). For simplicity of exposition, in this paper, we collectively refer to the right side of the figure as the \textit{Decoder}, while all components preceding it are referred to as the \textit{Encoder}.

To begin with, a new encoder was designed as the left side of Fig.\ref{fig.main}~(a). 
The core change of this encoder is that it receives visual prompts related to roofs. This allows PolyFootNet to be integrated with other popular near-nadir roof extraction methods with an external proposal model hub. Except for prompts from external models, the encoder includes a lightweight roof extraction module and an HQ mask prompter, which is composed of an HQ encoder and a segmentation decoder. The HQ encoder will enlarge the image embeddings from the image encoder in scale, and the segmentation decoder will decode them as auto roof prompts. The design of the HQ mask prompter referenced the design of Segmenter~\cite{segmenter}.
Then, the prompt encoder will encode them as prompt tokens. Based on the number of prompts, each prompt token will be allocated mask tokens, a vertex token, and an offset token. 

On the right side, after the process of two layers of two-way decoder~\cite{sam}, these tokens will be divided into two streams: vertex tokens will be used to extract key points, and offset tokens will be used to describe their positions to footprints. Via direct coordinate computing, the model can extract footprint key points. From key points to polygons, we adopt a mask-guided connecting strategy similar to the approach used in HiSup~\cite{hisup}, leveraging segmentation results to guide the polygon construction process.




To clarify our designs of PolyFootNet, it is broken down into three components:
self offset attention, multi-solution of BFE, and training settings. 

\subsection{Self offset attention (SOFA)}
\label{sec:2}

In prior experiments\cite{obm}, it was observed that models in prompt mode performed better when extracting footprints of taller buildings with significant offsets compared to lower buildings. 

In this paper, we propose a novel, trainable SOFA mechanism based on Nadaraya-Watson Regression~\cite{nadaraya, watson}, and Look-Ahead Masking~\cite{allattention} techniques used in Natural Language Processing (NLP). 
The SOFA module is designed to address the performance disparity seen in buildings with different offset lengths. 
The diagram for SOFA is presented in {{Fig.\ref{fig.main}}}.

In machine learning, attention layers can be interpreted as pooling mechanisms. 
The key role of SOFA is to aggregate offset information, particularly in cases where longer offsets are more reliable.

To introduce SOFA, we begin with Nadaraya-Watson Kernel Regression, a non-parametric regression method. 
Within SOFA, this technique is applied to accumulate offset knowledge, a vital aspect of the model's improved functionality. 
The Nadaraya-Watson Kernel Pooling can be described as follows:
\begin{equation}
  \label{eq:NWK}
 f(\mathbf{x}) = \sum_{i=1}^{n}\frac{\mathcal{K}(\mathbf{x}-x_i)}{\sum_{j=1}^{n}\mathcal{K}(\mathbf{x}-x_j)}y_i , \\
\end{equation}
where $f$ represents the Nadaraya-Watson Kernel Regression, and $\mathcal{K}$ denotes the kernel function. In the context of Transformers, $\mathbf{x}$ refers to the query, ${x_1, x_2, \dots, x_n}$ are the keys, and ${y_1, y_2, \dots, y_n}$ are the values.
This equation can be interpreted as a weighted sum of the values, where the weights are computed based on the similarity between the query ($\mathbf{x}$) and the keys (${x_1, x_2, \dots, x_n}$).This equation can be interpreted as a weighted sum of the values, where the weights are computed based on the similarity between the query ($\mathbf{x}$) and the keys (${x_1, x_2, \dots, x_n}$).
In essence, Nadaraya-Watson Kernel Regression operates analogously to an attention mechanism, where the kernel function $\mathcal{K}(\mathbf{x} - x_i)$ serves to compute a soft similarity between the query and each key. These similarities are then normalized to form an attention mask, which is subsequently used to perform a weighted average over the values.

In the BFE problem, our prior knowledge tells us that longer offsets tend to have better direction. 
Based on the interpretation of Kernel Regression, for determining the offset angle, we compute weights according to the relationship and similarity between each offset. Then, the weights will be used to  computing final angles for each offset. 
Therefore, Eq.\ref{eq:NWK} is translated as Eq.\ref{eq:sofa_1}:
\begin{equation}
  \label{eq:sofa_1}
 \boldsymbol{\dot\alpha} = f(\boldsymbol{\rho}, \boldsymbol{\alpha}) = \sum_{i=1}^{n}\frac{\mathcal{K}(\boldsymbol{\rho}-\rho_i)}{\sum_{j=1}^{n}\mathcal{K}(\boldsymbol{\rho}-\rho_j)}\alpha_i . \\
\end{equation}

In Eq.\ref{eq:sofa_1}, offset queries $\vec{\mathbf{O}} = \{{\mathbf{o_1}}, {\mathbf{o_2}}, ..., {\mathbf{o_n}}\}$ was expressed as polar coordinate $\vec{\mathbf{O}}=\{(\rho_1,\alpha_1), ...,(\rho_n,\alpha_n)\}$. 
$\boldsymbol{\rho}$ means length of the offset, and $\boldsymbol{{\alpha}}$ represents angle of the offset. $\boldsymbol{\dot{\alpha}}$ is the corrected offset angle. 

To facilitate computation, $\mathcal{K}$ is defined as Gaussian Kernal:
\begin{equation}
  \label{eq:sofa_2}
 \mathcal{K}(u) = \frac{1}{\sqrt{2\pi}}e^{(-\frac{u^2}{2})} .\\
\end{equation}
Consequently, Eq.\ref{eq:sofa_1} transformed as:
\begin{align}
   \label{eq:sofa_3}
 f(\boldsymbol{\rho},\boldsymbol{\alpha}) = \sum_{i=1}^{n}\textrm{softmax}\left(-\frac{1}{2}(\boldsymbol{\rho}-\rho_i)^2\right)\alpha_i .
\end{align}
To describe more complicated similarity, we make the whole attention learnable, and a trainable parameter $w$ was added to Eq.\ref{eq:sofa_3}:
\begin{align}
  \label{eq:sofa_4}
 \boldsymbol{\dot{\alpha}} &= \textrm{SOFA}_{a}(\boldsymbol{\rho}, \boldsymbol{\alpha}) \\ \nonumber
 &= \sum_{i=1}^{n}\textrm{softmax}\left(-\frac{1}{2}\left(w\times(\boldsymbol{\rho}-\rho_i)\right)^2\right)\alpha_i .
\end{align}

In NLP, look-ahead masking was used to combine the fact that tokens at one stage should only make the current and past knowledge visible for the model. 
Tokens in subsequent positions were masked via a substantial negative number in $\textrm{softmax}({\cdot})$. 
In the BFE problem, the longer offsets perform better than shorter offsets. 
Based on the aforementioned ideas, we need to design a look-longer masking $\boldsymbol{\mathcal{M}}$ for those shorter offsets. 
Finally, angle-level SOFA can be expressed as:

\begin{align}
  \label{eq:sofa_angle}
 \boldsymbol{\dot{\alpha}} &= \textrm{SOFA}_{a}(\boldsymbol{\rho}, \boldsymbol{\alpha}) \\ \nonumber
 &= \sum_{i=1}^{n}\textrm{softmax}\left(-\frac{1}{2}\left(w\times\boldsymbol{\mathcal{M}}(\boldsymbol{\rho}-\rho_i)\right)^2\right)\alpha_i .
\end{align}

Based on similar mathematical reasoning processes, vector-level SOFA can be written as:

\begin{align}
  \label{eq:sofa_vector}
 \mathbf{\dot{O}} &= \textrm{SOFA}_{v}(\boldsymbol{\rho}, \vec{\mathbf{U}})\\ \nonumber
 &= \boldsymbol{\rho}^\top\sum_{i=1}^{n}\textrm{softmax}\left(-\frac{1}{2}(w\times\boldsymbol{\mathcal{M}}(\boldsymbol{\rho}-\rho_i))^2\right)\mathbf{u_i} ,
\end{align}
$\vec{\mathbf{U}}=\{\mathbf{u_i}\}$ is the unit offset. $\mathbf{\dot{O}}=\{\mathbf{\dot{o_i}}\}$ is the output offset.

The $w$ among the mentioned operations was set to 0. 
Based on our experiment, this parameter fluctuated at around 0 while training. 

From Eq.\ref{eq:sofa_angle} and Eq.\ref{eq:sofa_vector}, SOFA is a portable and plug-and-play block, 
because (1) the learnable parameter $w$ was light; 
(2) in one off-nadir image \textit{I}, the number of buildings $N$ tends to be under 100, making the matrix's spatial operation not consume much GPU memory.

\subsection{Multi-solution of BFE}
\label{sec:raginbfe}
In the BFE problem, most of the studies are intuitively established based on the mask similarity between roofs and footprints, as shown in Fig.\ref{fig.intro}(a). 
Inspired by the third law of geography, which posits that
information in spatial representation is often redundant; we want to explore different solutions of BFE to determine whether \textit{roof + offset} is the only solution to the BFE problem. 
Under this setting, we explore the solution of Fig.\ref{fig.intro}(b) and (c).

In (c), the extracting method is also explicit: move the building mask along the roof-to-footprint offset, and the union between that and the building mask is the footprints. 
Similarly, if the building mask were moved in the opposite direction, the union would be the roof mask. 

In (b), the situation was more complicated. The first step was to find a direction to better represent the direction from roof to footprint. 
In this direction, a search algorithm was applied to detect the location of footprints by valuing different lengths of movement. 
Of course, we can also use the predicted global offset direction as this direction. 
In Algorithm \ref{alg:footprint_searching}, we introduced how to extract building footprint with only a roof and building segmentation.

\begin{algorithm}[!h]
	\caption{Footprint Searching}
    \label{alg:footprint_searching}
    \begin{algorithmic}[1]
      \REQUIRE Roof segmentation $r_i$, building segmentation $b_i$, maximum iteration $N$, offset step length $\bar{l}$
      \ENSURE Related footprint segmentation $f_i$
      \STATE Define angle set: ${\boldsymbol{\alpha}} \gets \{0^\circ, 1^\circ, \dots, 359^\circ\}$
      \STATE Initialize optimal angle $\theta \gets 0$, optimal offset length $l \gets 0$, and maximum overlap $S_{\text{max}} \gets 0$
      \FOR{$\alpha \in \boldsymbol{\alpha}$}
          \STATE Construct movement matrix:
          \[
          \mathbf{V}_\alpha = \begin{bmatrix}
            1 & 0 & -\bar{l}\cos{\alpha}\\[2pt]
            0 & 1 & -\bar{l}\sin{\alpha}\\[2pt]
            0 & 0 & 1
          \end{bmatrix}
          \]
          \STATE Move roof segmentation: $\dot{R}_\alpha \gets \mathbf{V}_\alpha \cdot r_i$
          \STATE Calculate overlap ratio:
          \[
            S_\alpha \gets \frac{\mathrm{Area}(\dot{R}_\alpha \cap b_i)}{\mathrm{Area}(r_i)}
          \]
          \IF{$S_\alpha > S_{\text{max}}$}
              \STATE Update optimal angle: $\theta \gets \alpha$
              \STATE Update maximum overlap: $S_{\text{max}} \gets S_\alpha$
          \ENDIF
      \ENDFOR
      \STATE Determine optimal offset length:
      \[
        l \gets \underset{l_c}{\arg\min}\left|\frac{\mathrm{Area}(\mathbf{V}_{\theta,l_c}\cdot r_i \cap b_i)}{\mathrm{Area}(r_i)} - 1\right|
      \]
      where
      \[
        \mathbf{V}_{\theta,l_c} = \begin{bmatrix}
        1 & 0 & -l_c\cos{\theta}\\[2pt]
        0 & 1 & -l_c\sin{\theta}\\[2pt]
        0 & 0 & 1
        \end{bmatrix}
      \]
      \STATE Compute final footprint segmentation:
      \[
      f_i \gets \mathbf{V}_{\theta,l}\cdot r_i
      \]
    \end{algorithmic}
\end{algorithm}

From lines 1 to 11 in Algorithm~\ref{alg:footprint_searching}, a linear search was applied to identify the optimal rotation angle for moving the roof segmentation to best overlap with the corresponding building segmentation. Subsequently, given this optimal angle, a binary search was utilized to determine the optimal offset length along this angle precisely.

Specifically, when employing spatial information in Algorithm~\ref{alg:footprint_searching} to extract building footprints, the optimal offset length is determined based on the critical condition illustrated in Fig.\ref{fig.edge}. In the algorithm, the expression $\mathbf{V}_{\theta,l}\cdot r_i$ denotes the spatial translation of the roof mask $r_i$ along the optimal direction $\theta$ by the offset length $l$, where $\mathbf{V}_{\theta,l}$ represents the corresponding homogeneous movement matrix.

Fig.\ref{fig.edge}(c) demonstrates the method for accurately extracting offsets. As shown, the binary search mentioned previously was conducted twice: once to determine the optimal {\color{yellow}{yellow offset}} and once for the optimal {\color{red}{red offset}}. Given that the {\color{blue}{blue offset}} length equals the {\color{yellow}{yellow offset}}, the final roof-to-building offset is calculated as the difference between the lengths of the {\color{red}{red offset}} and the {\color{yellow}{yellow offset}}.

\subsection{Training settings}
The OBM's losses were combined with two parts: prompt-level segmentation loss and offset losses in ROAM, which can be expressed as:
\begin{equation}
\mathcal{L}_{OBM} = \mathcal{L}_{ROAM} + \mathcal{L}_{roof} + \mathcal{L}_{building},
\label{eq:obmloss}
\end{equation}
where \( \mathcal{L}_{ROAM} \) is the loss of ROAM, which applies SmoothL1 Loss~\cite{fastrcnn} for each offset head. 
\( \mathcal{L}_{roof} \) is CrossEntropy Loss~\cite{celoss} of roof segmentation, and \( \mathcal{L}_{building} \) is CrossEntropy Loss of building segmentation.

For two new tasks, CrossEntropy Loss is applied for roof semantic segmentation. The model outputs a vertex map for each building in prompt-level vertex segmentation. 
Because the vertex map contains the whole scope of the inputted image, most pixels will be valued as a negative sample by 0. 
Sparse positive key points will mislead the model to predict negative samples only. 
Via experiments, if the model uses a fixed window size to crop the building area, it can lead to severe grid effects. 
As a result, DS-BCE Loss was designed for prompt-level vertex segmentation:
\begin{equation}
 \mathcal{L}_{vetex} = \sum^{p\in Z+\Delta}{-y_p \log{y'_p} -(1-y_p)\log(1-y'_p)},
 \label{eq:4}
\end{equation}
$y_p$ and $y'_p$ are pixels on ground truth and prediction maps at $p$. 
$Z$ is the original prompted area, and $\Delta$ is a random small neighborhood of this area.
Finally, PolyFootNet was trained as:
\begin{equation}
 \mathcal{L} = \lambda \times (\mathcal{L}_{OBM} + \beta\mathcal{L}_{vertex}) + \kappa\times\mathcal{L}_{seg},
 \label{eq:obmlossv2}
\end{equation}
$\lambda$ and $\kappa$ were parameters equal to 1 or 0 to control whether semantic heads would be trained together. 
$\beta$ is a scale factor to balance the losses of vertex tasks and other tasks. 

%% file: sec/3_exp.tex
\input{chapters/Exp}

%% file: chapters/Exp.tex
\section{Experiment and analysis}
\label{sec:Exp}

\begin{table*}[t]
    \caption{Main results on BONAI \cite{BONAI}.}
    \label{tab:main}\resizebox{\linewidth}{!}{
    \setlength{\arrayrulewidth}{1.05 pt}
    \renewcommand{\arraystretch}{1.1}
    \begin{tabular}{lcccccccccc}
    \hline                                                
    Model & F1&	\small{Precision}&	Recall	&EPE	&m$VE$	&m$LE$	&m$AE$	&a$VE$	&a$LE$	&a$AE$	\\\hline
    PANet& 58.06&59.26&56.91&-&-&-&-&-&-&-\\
    M RCNN&58.12&59.26&57.03&-&-&-&-&-&-&-\\
    MTBR-Net&63.60&64.34&62.87&5.69&-&-&-&-&-&-\\
    LOFT&64.42&64.43&64.41&4.85&-&-&-&-&-&-\\
    Cas.LOFT$^\star$&62.58&63.67&61.52&4.79&-&-&-&-&-&-\\
    MLS-BRN&66.36&65.90&66.83&4.76&-&-&-&-&-&-\\\hline
    p.LOFT    &72.98&85.74&64.01&-&15.4&\textbf{12.6}&0.18&6.12&4.51&0.32\\
    p.Cas.LOFT$^\star$&76.05&\textbf{87.20}&67.82&-&15.8&13.5&0.17&5.97&4.48&0.31\\
    OBM      &\textbf{80.03}&82.57&\textbf{77.97}&-&\textbf{15.3}&13.9&0.12&\textbf{5.12}&\textbf{4.05}&0.22\\
    OBM{$^\dag$}      &78.65&80.41&77.21&-&17.0&15.7&0.12&5.38&4.35&0.22\\\hline
    \small{Ours(mask)}&78.74&79.85&77.91&\multirow{2}{*}{-}&\multirow{2}{*}{17.0}&\multirow{2}{*}{15.8}&\multirow{2}{*}{\textbf{0.11}}&\multirow{2}{*}{5.46}&\multirow{2}{*}{4.40}&\multirow{2}{*}{\textbf{0.22}}\\
    \small{Ours(poly.)}&75.31&78.12&73.13\\
    \hline
    \end{tabular}}
\end{table*}

This section describes the data used in this work, justifies their choice, and specifies their sources.
Then, the main results of the models are reported and analyzed.
Lastly, a generalization test was conducted on the Huizhou test set. 

\subsection{Dataset}
In our experiments, three datasets were employed to make a comparison: 

\textbf{BONAI}~\cite{BONAI}: This dataset was launched with benchmark model LOFT. 
There are 3,000 train images and 300 test images, and the height and width of the images are 1024. 
Building annotations include roof segmentation, footprint segmentation, offset, and height.

\textbf{OmniCity-view3}~\cite{omnicity}: OmniCity-view3 provided 17,092 train-val images and 4,929 test images in height and width 512. 
Footprint segmentation, building height, roof segmentation, and roof-to-footprint offset were labeled for each building. 

\textbf{Huizhou test set}~\cite{obm}: This small dataset labeled images from a new city Huizhou, China. 
All buildings were plotted point-to-point by human annotators. 
The shape of the images is the same as that of BONAI, and there are over 7,000 buildings with offsets, roofs, and footprint segmentation. 

In comparison, OmniCity images have a relatively higher spatial resolution and smaller cropped image sizes than those from BONAI and Huizhou.
\subsection{Metrics}
\subsubsection{Offset metrics}
Basically, we measure each offset in three aspects: Vector Error ({$VE$}), Length Error (\textit{$LE$}), and Angle Error ($AE$), which defined as:
\begin{equation}
  VE = \left|\vec p - \vec p_g \right|_2;\\
  LE = \big| \left|\vec p\right|_2 - \left|\vec p_g \right|_2\big|;\\
  AE = |\theta - \theta_g|,
\end{equation}
where \( \vec p \) and \( \vec p_g \) represent the predicted and ground truth offset.
\( \theta \) and \( \theta_g \) represent the predicted and ground truth angle.
The \( \left|\cdot\right| \) and \( \left|\cdot\right|_2 \) represent the 1-Norm and 2-Norm respectively.

Referring to COCO metrics~\cite{coco}, each offset will be grouped based on the length of its related ground truth. 
\begin{equation}
   m\mathcal{E} = \frac{\mathcal{E}_{(10n,\infty)} + \sum_{i=0}^{n} \mathcal{E}_{(10i,10i+10)}}{n+1} \\
\end{equation}
\noindent where \(\mathcal{E}\) can represent $VE$, $LE$ and $AE$.
 \(\mathcal{E}_{(i,j)}\) represents the mean error of the offset group whose length is between $i$ and $j$. 
Specifically, \(\mathcal{E}_{(0,\infty)}\) is named as average error $a\mathcal{E}$.

For instance, in segmentation-based methods, we compute the object-wise end-point error~\cite{BONAI} in pixels to evaluate the Euclidean distance between the endpoints of the predicted and ground truth offset vector.

\subsubsection{Mask metrics}
Precision, Recall and F1score is used to evaluate the quality of footprint masks and polygons.

{\subsection{Baseline introduction}
To demonstrate the effectiveness of our method, the results of the following models were chosen as comparative experiments.
\begin{itemize}
    \item Path Aggregation Network(PANet)~\cite{panet} is an instance segmentation model.
    \item Mask RCNN~\cite{maskrcnn} is an ROI-based instance segmentation network. 
    \item MTBR-Net~\cite{MTBRNet} is a semantic segmentation network based on global offset features used for extracting building footprints. 
    \item LOFT~\cite{BONAI} is an "instance-segmentation-based" footprint extraction model. 
    \item MLS-BRN~\cite{MLSBRN} is a multi-task model that gives more diverse building-related information, like shooting angles and building height.
    \item OBM\cite{obm} is a specially designed model for promptable footprint extraction. 
    
\end{itemize}}

\subsection{Experimental settings}
In the training process of our PolyFootNet was trained in two stages.
All images will be reshaped in 1024$\times$1024 pixels.
Experiments were conducted on a server with 4 NVIDIA RTX 3090.
In the first stage, PolyFootNet was trained in roof prompting mode. Then, proposal networks were trained solely.
In the whole training process, we used stochastic gradient descent (SGD) \cite{sgd} as the optimizer with a batch size of 4 for 48 epochs, 
an initial step learning rate of 0.0025, a momentum of 0.9, and a weight decay of $10^{-4}$.
The number of parameters in PolyFootNet is 77.69 M. 

{In this paper, the experiments will be divided into three groups: in the experiments on the datasets BONAI and OmniCity, we will use the author's original training and testing set partitioning criteria, respectively. For the experiments on the Huizhou testing set, the models trained in the BONAI experiment will be directly tested.}

\subsection{Main results}
In this section, we evaluate our methods on BONAI and OmniCity-view3. 
The primary comparison will focus on LOFT \cite{BONAI}, Cascade LOFT and OBM between PolyFootNet because these models are available for extensive experiments.
    
In the listed tables, the displayed results were separated into three parts. 
The first part lists results from end-to-end models. 
Note that: $^\star$ model is a model that we reproduce based on the author's intention in paper~\cite{BONAI}.OBM{$^\dag$} was retrained under the same training setting as PolyFootNet. 
p. is short for \textit{prompt}. 
Additionally, EPE, m$VE$, m$LE$, a$VE$, and a$LE$ were in pixels. Precision, Recall, F1score were measured in percentage(\%). 

\begin{figure*}[!h]
	\centering
	\includegraphics[width=\linewidth]{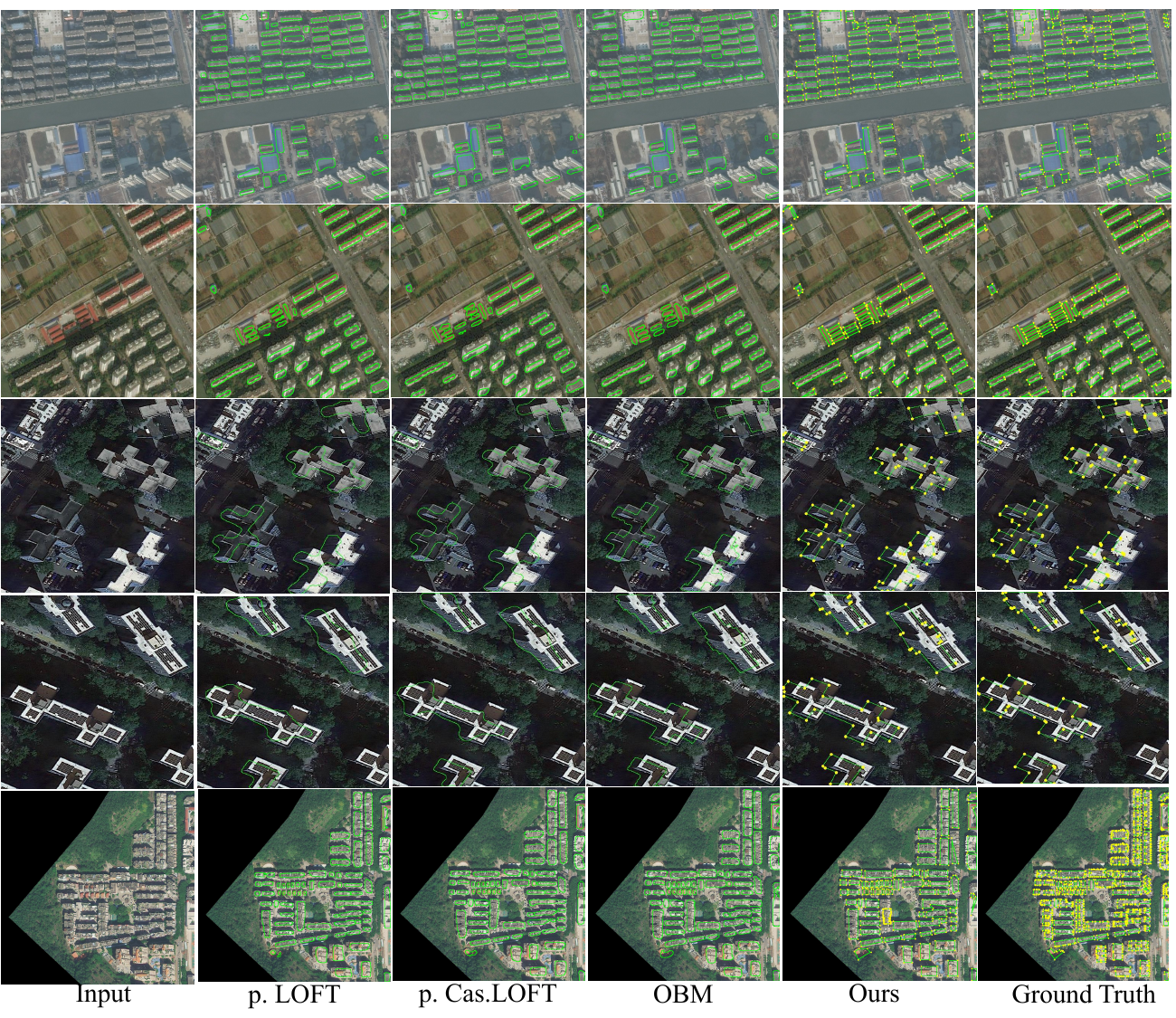}
	\caption{Main results extracted by prompting mode. The green lines represent predicted building footprint boundaries, and the yellow points are key nodes of the building footprints.}
	\label{fig.mainresults}
\end{figure*}

All available promptable models were compared with our proposed model. 
From Tab.\ref{tab:main}, although PolyFootNet can provide better roofs, the quality of offsets and footprints is not as good as the formerly proposed methods.  
Additionally, except experiments on OmniCity in Tab.\ref{tab:omni}, clear drops between polygon results and mask results was found on Huizhou and BONAI datasets: their F1score dropped by 3.43\% and 4.01\%. 
This may be caused by the resolution of images. As shown in Fig.\ref{fig.mainresults}, BONAI and Huizhou datasets have lower pixel resolution compared with OmniCity dataset, leading to less detailed edge information of buildings. Consequently, the key point extraction and point connections are hard to do compared with those on OmniCity. 
Extensive experiments were conducted in ablation studies to examine this hypothesis. 

Experiments on OmniCity-view3 demonstrate the advance of our PolyFootNet. In Tab.\ref{tab:omni}, PolyFootNet nearly outperformed all mentioned models. 
\begin{table*}[!h]
    \caption{Experimental results on OmniCity-view3.}
    \label{tab:omni}\resizebox{\linewidth}{!}{
    \setlength{\arrayrulewidth}{1.05 pt}
    \renewcommand{\arraystretch}{1.1}
    \begin{tabular}{@{\extracolsep{\fill}}lcccccccccc}
    \hline                                                
    Model &F1 &Precision &Recall &EPE	&m$VE$	&m$LE$	&m$AE$	&a$VE$	&a$LE$	&a$AE$\\\hline
    M RCNN&69.75&69.74&69.76&-&-&-&-&-&-&-\\
    LOFT&70.46&68.77&72.23&6.08&-&-&-&-&-&-\\
    MLS-BRN&72.25&69.57&75.14&5.38&-&-&-&-&-&-\\\hline
    p.LOFT    &82.27&90.63&75.81&-&54.3&48.5&0.65&7.57&5.29&0.70\\
    p.Cas.LOFT&83.75&\textbf{91.62}&77.54&-&52.9&48.4&\textbf{0.62}&7.25&5.12&0.69\\
    OBM&86.03&90.17&82.52&-&56.9&53.7&0.66&6.69&5.15&0.64\\ \hline
    Ours(mask)&\textbf{88.42}&90.06&\textbf{87.01}&\multirow{2}{*}{-}&\multirow{2}{*}{\textbf{51.5}}&\multirow{2}{*}{\textbf{47.9}}&\multirow{2}{*}{0.63}&\multirow{2}{*}{\textbf{6.15}}&\multirow{2}{*}{\textbf{4.65}}&\multirow{2}{*}{\textbf{0.59}}\\
    Ours(poly.)&87.61&90.76&84.93    \\\hline
    \end{tabular}}
\end{table*}
Although vectorized footprint polygons still have lower F1score than their mask footprints by 0.81\%, PolyFootNet outperformed all mentioned models. 
In PolyFootNet, polygonal results even have better Precision than mask results(+0.7\%). 
Moreover, PolyFootNet performed better than any other model in terms of offset prediction. 
\textit{a$VE$} of PolyFootNet is 1.423 pixels lower than that of prompt LOFT and 0.544 pixels lower than the figure for OBM. 

In Fig.\ref{fig.mainresults}, visualized results in the prompt mode were provided. 
The first and second lines of illustrations were selected from BONAI~\cite{BONAI}. 
The third and fourth lines were from OmniCity-view3~\cite{omnicity}, and the last line was from Huizhou~\cite{obm}. 
Our model can provide more editable polygonal results compared with other models.

A generalization test was conducted at the Huizhou test set in Tab.\ref{tab:huizhou}.
All models were pre-trained on BONAI, and there was no extra training. 
In predicting footprints, results of OBM and PolyFootNet exhibit similar attributes with them on BONAI, 
\textit{e.g.} the gap between F1scores of footprints predicted by PolyFootNet on BONAI (75.31\%) and Huizhou Test (75.35\%) is 0.04\%, and a similar gap between mask and polygon results were found. 

In summary, PolyFootNet can directly predict footprint polygons, the performance of which showcased a certain generalization ability.

\subsection{Multi-solution for BEF problem}
Multi-information can be divided into two classes: information extracted by different kinds of building-related models and extracted multi-information by PolyFootNet. 
Human-plotted prompts will motivate models to reach their ceiling performance. 
In this part, footprints F1score were selected to measure mask ability; meanwhile, EPE and a$VE$, which are similar in definition, were used to measure offset ability. 
Each model will be scattered on coordinates. 
To facilitate comparison, BONAI was used due to its diverse open-source methods and known experimental results.
\begin{figure}[!h]
	\centering
	\includegraphics[width=1\linewidth]{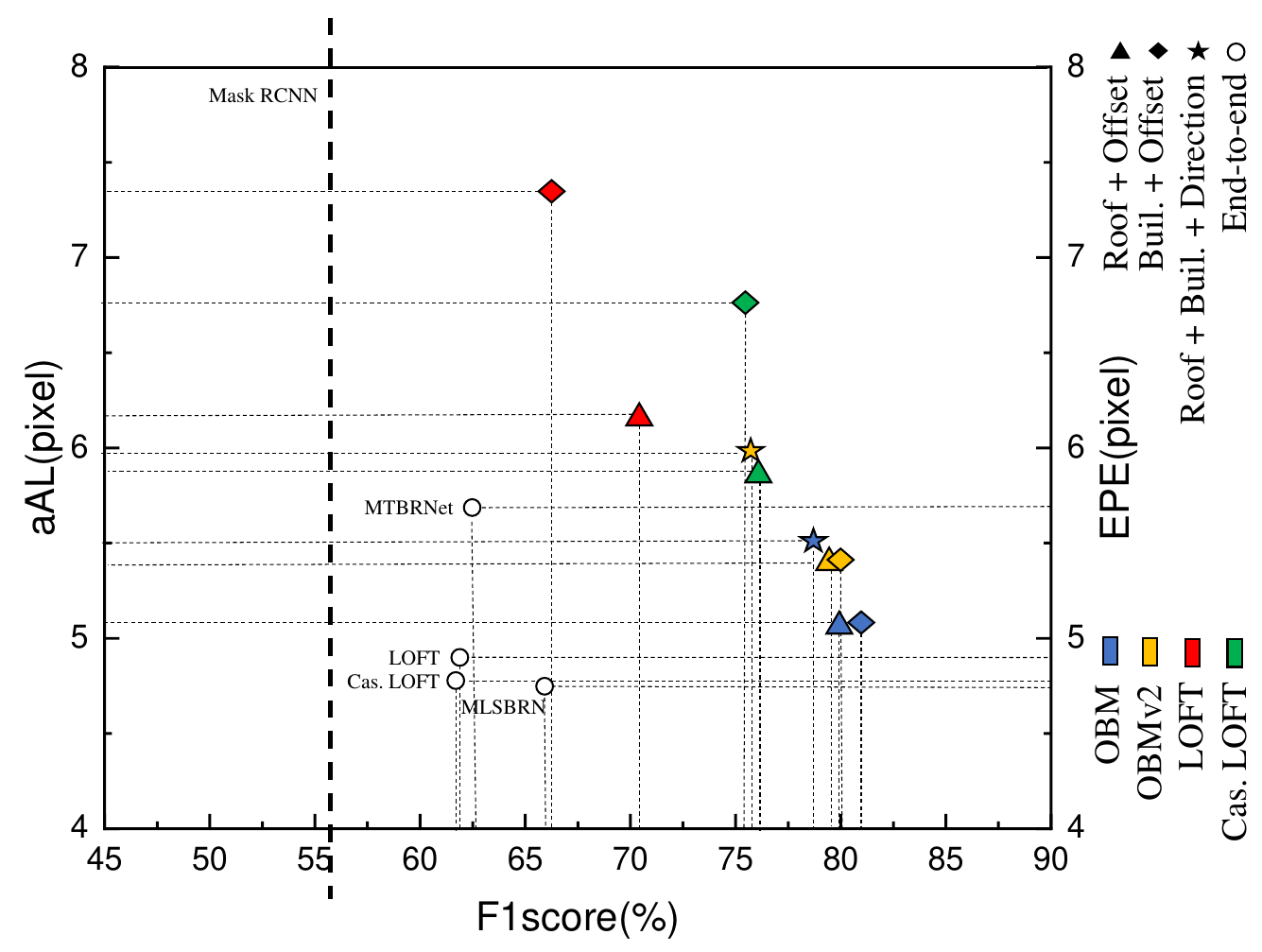}
	\caption{Extracting footprints with multi-solutions of BFE.}
	\label{fig.multiSource}
\end{figure}

In Fig.\ref{fig.multiSource}, the BFE problem was solved with different information. 
For promptable models, OBM and PolyFootNet, building segmentation with offsets can predict better building footprint (F1score +1.37 and +0.23 respectively). 
End-to-end ROI-based can also extract footprint with building segmentation and offsets, which provides similar performance with related models using roof and offset. 
Additionally, extracting footprints with roof and building segmentation has been proved applicable, although offsets regressed in this version were inaccurate compared with models using roofs. 
All models can provide better results than Mask RCNN.

The automatic extraction of building footprints commonly relies on proposal regions. 
In Tab.\ref{tab:auto}, PolyFootNet was tested with different region proposal functions. 
Additionally, utilizing different sources of information to extract footprint print was also examined. 

r., b., o. and d. represent roof mask, building mask, offset and offset direction. 

With the help of HTC, PolyFootNet can automatically extract building footprints, 
and the Recall of this model is higher (+8.16) than the former SOTA MLS-BRN. 
By adjusting different prompting modes, PolyFootNet can also reach a similar precision to MLS-BRN (-1.24).

\begin{figure}[!h]
	\centering
	\includegraphics[width=0.9\linewidth]{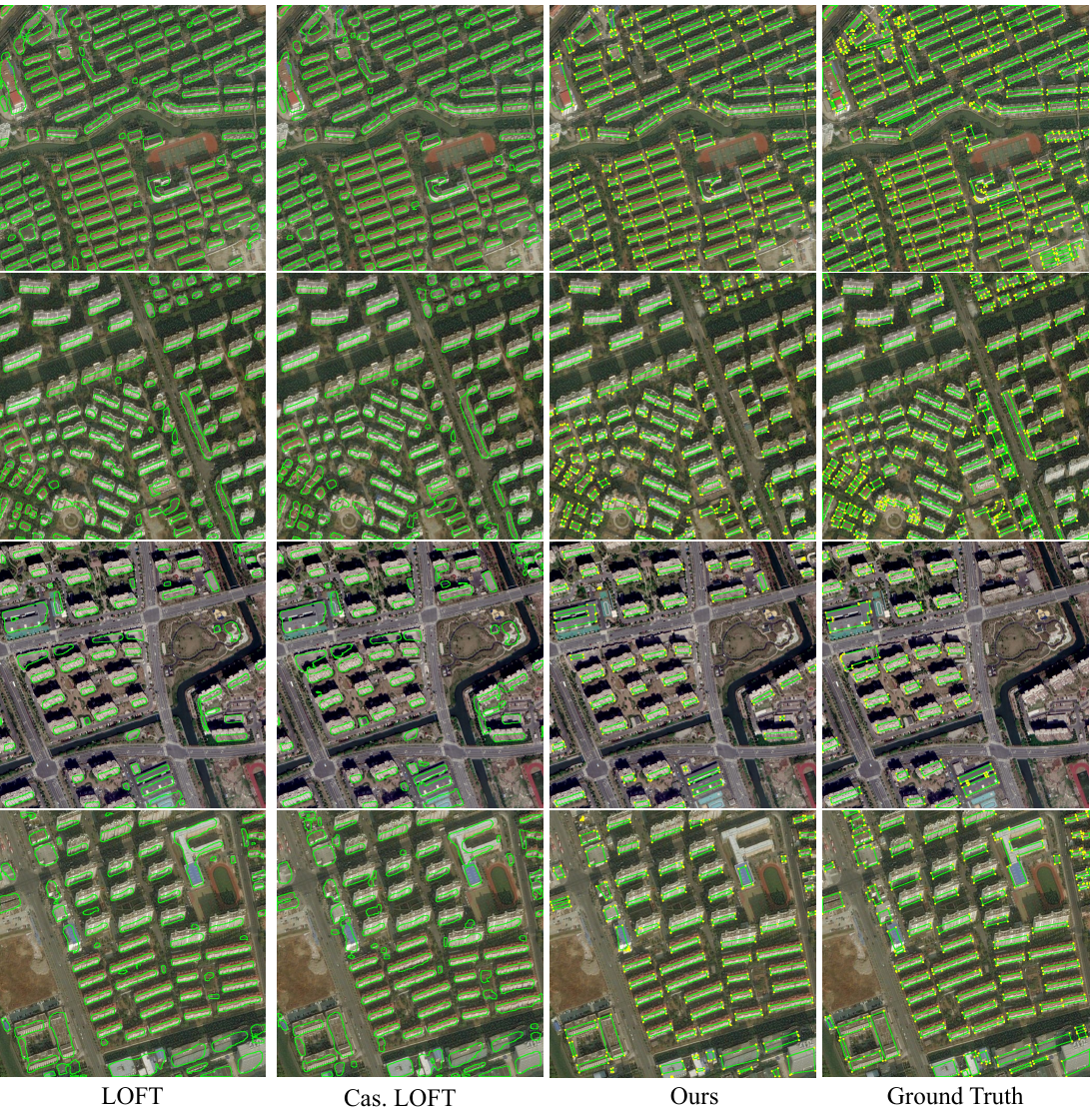}
	\caption{Main results extracted by auto mode}
	\label{fig.autoresults}
\end{figure}

In Fig.\ref{fig.autoresults}, footprints extracted by auto mode were compared with each other. 
Our model can still provide polygonal results compared with other models.  

\begin{table*}[htbp]
    \centering
    \begin{minipage}{0.48\textwidth} 
        \centering
    \caption{Experimental results on Huizhou test set.}
    \label{tab:huizhou}\resizebox{\linewidth}{!}{
    \setlength{\arrayrulewidth}{1.05 pt}
    \renewcommand{\arraystretch}{1.1}
    \begin{tabular}{@{\extracolsep{\fill}}lcccccc}
    \hline                                                
    Model &F1 &Precision &Recall&a$VE$	&a$LE$	&a$AE$\\\hline
    p.LOFT	&72.56	&83.02	&65.33	&7.935	&6.449	&0.752\\
    p.Cas.LOFT	&75.83	&81.71	&71.13	&7.894	&5.938	&0.818\\
    OBM	& 81.53	&78.80	&84.80	&4.898	&4.351	&0.169\\
    OBM{$^\dag$}&80.30&79.04&81.85&4.985&4.412&0.198\\\hline
    Ours(mask)	&80.36	&78.99	&82.18	&\multirow{2}{*}{4.959}&\multirow{2}{*}{4.636}&\multirow{2}{*}{0.144}\\
    Ours(poly.)	&75.35	&77.04	&74.25	\\\hline
    \end{tabular}}
    \end{minipage}
    \hfill
    \begin{minipage}{0.48\textwidth} 
        \centering
    \caption{Auto extraction results on BONAI test set.}
    \label{tab:auto}
    \resizebox{\linewidth}{!}{
    \setlength{\arrayrulewidth}{1.05 pt}
    \renewcommand{\arraystretch}{1.1}
    \begin{tabular}{@{\extracolsep{\fill}}lcccc}
    \hline                                                
    Model &F1 &Precision &Recall &EPE\\\hline
    M RCNN &56.12	&57.02	&55.26&-\\
    LOFT(r.o.)	&64.42	&64.43	&64.41	&4.85\\
    LOFT(b.o.) &61.97	&61.41	&62.55 &5.83\\
    Cas.LOFT(r.o.)	&62.58	&63.67	&61.52&4.79    \\
    Cas.LOFT(b.o.)&61.67	&64.59	&59.00 &5.23\\
    MLS-BRN&\textbf{66.36}&\textbf{65.90}&66.83&4.76\\\hline
    Ours+HTC(r.o.)&61.48	&55.19	&74.79&\multirow{2}{*}{\textbf{4.59}}\\
    Ours+HTC(b.o.)&61.48 &55.13&\textbf{74.99}\\\hline
    Ours+seg.&55.20&64.66&50.47&-\\\hline
    \end{tabular}
    }
    
    \end{minipage}

\end{table*}

%% file: sec/4_ablation.tex
\input{chapters/Ablation}

%% file: chapters/Ablation.tex
\section{Ablation studies}
\label{sec:Abl}
This section will examine the proposed algorithms and modules. 
Apart from the SOFA module, the aforementioned algorithms proposed for extracting footprints with different information also need ablation. 

\subsection{Pixel resolution caused marginal performance}
Polygonal footprints are typically generated by connecting the extracted key points based on the orientation of the mask footprint contours. During this process, discrepancies between mask results and polygon results are inevitable. However, compared to the OmniCity dataset, the performance discrepancies observed on the BONAI and Huizhou datasets are significantly greater. As shown in Fig.\ref{fig.mainresults}, the spatial resolution of most images in OmniCity is considerably higher than that of BONAI and Huizhou datasets. This leads to buildings occupying more effective pixels within images, providing clearer key points and edges, thereby reducing the differences between vectorization and mask results.

To validate this hypothesis, we artificially simulated higher resolutions by upsampling images from the BONAI and Huizhou datasets, splitting each image evenly into four sub-images, and then retrained and retested the model. In Tab.\ref{tab:upsampled}, our model outperformed OBM, and the performance gap between mask and polygon results narrowed compared with those on original datasets.

\begin{table}
    \centering
    \caption{Performance comparison under upsampled BONAI and Huizhou dataset}
    \label{tab:upsampled}\resizebox{\linewidth}{!}{
    \setlength{\arrayrulewidth}{1.05 pt}
    \renewcommand{\arraystretch}{1.1}
    \begin{tabular}{lcccc}
        \hline   
         Model&  Dataset&  F1&  Precision& Recall\\\hline
         OBM&  \multirow{3}{*}{BONAI}&  63.18& 74.91& 55.41\\
         Ours(mask)&  &  74.97&  77.12& 73.56\\
         Ours(poly.)&  &  74.49&  76.89& 73.00\\\hline
         OBM&  \multirow{3}{*}{Huizhou}&  57.99&  73.94& 48.61\\
         Ours(mask)&  &  72.94&  73.96& 72.61\\
         Ours(poly.)&  &  72.73&  73.15& 73.09\\\hline
    \end{tabular}}

\end{table}
\subsection{Ablation of SOFA}
Unlike other transformer modules, SOFA was developed not sensitive to the input length of embedded tokens. 
Because of this feature, SOFA module can adapt to any offset-based model. 
After all other modules finished training, the SOFA module was trained in the last stage. 
In Tab.\ref{tab:sofa}, SOFA module was applied in all open-source models and can reduce both prompt-level and instance-level offset errors.
\textit{e.g.} EPE of LOFT on the BONAI dataset declined by 0.33 pixels. 

\begin{table}[!h]
    \caption{SOFA module ablation studies on BONAI.}
    \label{tab:sofa}\resizebox{\linewidth}{!}{
    \setlength{\arrayrulewidth}{1.05 pt}
    \renewcommand{\arraystretch}{1.1}
    \begin{tabular}{lccccccc}
    \hline                                                
    Model &EPE	&m$VE$	&m$LE$	&m$AE$	&a$VE$	&a$LE$	&a$AE$\\\hline
    LOFT      &4.85&15.4&12.6&0.18&6.12&4.51&0.32\\
    LOFT+SOFA &\textbf{4.52}&\textbf{14.6}&\textbf{12.6}&\textbf{0.13}&\textbf{5.62}&\textbf{4.49}&\textbf{0.22}\\\hline
    Cas.LOFT  &4.79&15.8&13.5&0.17&5.97&4.48&0.31\\
    Cas.LOFT+SOFA  &\textbf{4.43}&\textbf{15.3}&\textbf{13.4}&\textbf{0.12}&\textbf{5.64}&\textbf{4.44}&\textbf{0.22}\\\hline
    OBM         &-&15.3&13.9&0.12&5.12&4.05&0.22\\
    OBM+SOFA &-&\textbf{15.3}&\textbf{13.9}&\textbf{0.11}&\textbf{5.08}&\textbf{4.04}&\textbf{0.21}\\ \hline
    OBM{$^\dag$}&-&17.0&15.7&0.12&5.38&4.35&0.22\\
    OBM{$^\dag$}+SOFA&-&16.8&15.7&0.11&5.36&4.35&0.21\\\bottomrule
    \end{tabular}}
\end{table}

\subsection{Multi-solutions of BFE}
Ablation studies were conducted with ground truth labels to ensure the proposed algorithms can extract building footprints. 

r., b., o. and d. represent roof mask, building mask, offset and offset direction. 

From Tab.\ref{tab:gt}, our proposed algorithms can accurately extract building footprints via different information. 
Although our algorithms can extract footprints merely with roof and building segmentations, 
grid-based digital images are always limited by image continuity. 
The influence of this problem is more severe, especially on the representations of buildings with short offsets, 
\textit{e.g.} an offset with length 1 pixel, there are only 4 points nearby that can represent its endpoint. 
This ambiguity leads to a poor perception of direction. 
As a result, when the offset direction was given, the $aVL$ dropped by 5.16 pixels and the f1-score of footprints increased by 6.5\%. 

\subsection{Proposal methods}
PolyFootNet can receive almost all kinds of models that can provide bounding boxes related to buildings. 
Except for an extra segmentation head on PolyFootNet, 
PolyFootNet was integrated with other models that can provide roof-bounding boxes or segmentations. 
These outputs perform as rough extraction results, and PolyFootNet will correct and refine them.

In Tab.\ref{tab:extr}, PolyFootNet was integrated with its segmentation head, HTC and LOFT. 
Specifically, HTC and LOFT are matched with different NMS strategies. 
$\spadesuit$ represents soft NMS algorithms with a score threshold of 0.05, IoU threshold of 0.5, and a maximum of 2000 instances per image. 
$\clubsuit$ represents NMS algorithms with a score threshold of 0.1, an IoU threshold of 0.5, and a maximum of 2000 instances per image. 
$\heartsuit$ leverages result from $\spadesuit$, but the same instance, which was repeatedly predicted, will be merged as one instance. 

Soft NMS algorithm gives almost all bounding boxes with a very low score threshold. 
This means most output boxes will be selected as final outputs. 
Consequently, they can provide results with high Recall, but PolyFootNet was trained with annotations that can cover the whole building or roof. 
As a result, the Precision was not good. 
\textit{e.g.} Precision of PolyFootNet + HTC$\spadesuit$ is lower than that of PolyFootNet + HTC$\heartsuit$ by 15.96\%, 
although the Recall of PolyFootNet + HTC$\spadesuit$ is higher than that of PolyFootNet + HTC$\heartsuit$ by 10.49\%. 
Finally, the score of PolyFootNet + HTC$\spadesuit$ was adversely influenced, which only reached 51.79\%. 

%% file: sec/5_discussions.tex
\input{chapters/Discussion}

%% file: chapters/Discussion.tex
\section{Discussion}
\begin{table*}
    \begin{minipage}{0.48\textwidth}
     \centering
    \hfill
    \caption{Extract footprints with different ground truth labels on BONAI test set.}
    \label{tab:gt}
    \resizebox{\linewidth}{!}{
    \setlength{\arrayrulewidth}{1.0 pt}
    \renewcommand{\arraystretch}{1.0}
    \begin{tabular}{@{\extracolsep{\fill}}lcccc}
    \hline                                                
     Model & F1 &Precision &Recall &a$VE$\\\hline
    b.+o.&98.22	&98.60	&97.88&0\\
    r.+o.	&98.55&99.30	&97.84	&0\\
    b.+r. &87.87	&88.67	&87.11 &5.83\\
    b.+r.+d.	&94.37&98.17	&91.69&0.67   \\\hline
    \end{tabular}}
    \end{minipage}
    \hfill
    \begin{minipage}{0.48\textwidth}
     \centering
    
    \caption{Extract footprints with other roof extraction models on BONAI test set.}
    \label{tab:extr}
    \resizebox{\linewidth}{!}{
    \setlength{\arrayrulewidth}{1.0 pt}
    \renewcommand{\arraystretch}{1.0}
    \begin{tabular}{@{\extracolsep{\fill}}lcccc}
    \hline                                                
    Model & F1 &Precision &Recall &EPE\\\hline
    PolyFootNet&78.74 &79.85 &77.91&-\\\hline
    +eve&59.67&68.66	&55.32	&5.03\\\hline
    +HTC$\heartsuit$&61.48&55.19	&74.80&4.59\\
    +HTC$\spadesuit$&51.79&39.23	&85.29&4.92 \\
    +HTC$\clubsuit$&60.70&59.18&66.32&4.42\\\hline
    +LOFT$\spadesuit$&56.24&44.01&84.33&5.07\\
    +LOFT$\clubsuit$&60.26&56.32&68.48&4.59\\\hline
    \end{tabular}
    }
    \end{minipage}
\end{table*}

\subsection{Try to understand SOFA}
The design of SOFA is inspired by prior knowledge: when predicting longer offsets, the model can provide relatively accurate directions but less precise lengths; whereas, for shorter offsets, the model tends to predict more accurate lengths but less precise directions. As shown in Eq.\ref{eq:sofa_vector}, SOFA introduces the concept of kernel regression in its design, determining the angular mixture weighting ratios by calculating the length relationships among all building offsets: when longer offsets come across shorter offsets, their related similarity will be larger than others. Combined with modern deep learning module design principles, this approach enhances the performance of building offset prediction at the module level.

To demonstrate the effectiveness of the SOFA module, Tab.\ref{tab:ip} presents the improvements in both length and angle prediction across different offset lengths after applying the SOFA vector version.
This table indicates that SOFA helps improve angle prediction for shorter offsets, as well as achieve more accurate overall offset predictions.

\begin{table}
    \centering
        \caption{Offset prediction improvement for different length groups with SOFA.}
    \label{tab:ip}
        \resizebox{\linewidth}{!}{
    \setlength{\arrayrulewidth}{1.0 pt}
    \renewcommand{\arraystretch}{1.0}
    \begin{tabular}{llrrrrr}\hline
        Model & $\mathcal{E}_{(i,j)}$ &  $(0,10)$&  $(10,20)$&  $(20,30)$&  $(80,90)$& $(90,100)$\\\hline
        Ours &  $VE$&  4.46&  4.31&  6.58&  23.01& 30.24\\
         (w/o)&  $LE$&   3.57&  3.10&  5.19& 21.17 &28.36 \\
         &  $AE$&  0.41&  0.17&  0.16&  0.09& 0.13\\\hline
        Ours &  $VE$&  3.91&  4.00&  5.92&  17.81& 23.19\\
         (w)&  $LE$&  3.20&  2.73&  4.46&  16.98& 22.22\\
         &  $AE$&  0.34&  0.16&  0.15&   0.04& 0.07\\\hline
    \end{tabular}}

\end{table}

\subsection{External models and multi-solutions of BFE}
The use of external prompts for interactive models has been studied in many cases.
\textit{e.g.} contrastive Language-Image Pre-training (CLIP) was trained on over 400 million pairs of images and text \cite{clip}. 
When researchers conduct experiments on classifying ImageNet \cite{ImageNet}. 
The researchers found that using a prompt template "A photo of a \{\textit{label}\}" can directly improve the accuracy by 1.3\% compared to using a single category word "\{\textit{label}\}". 
In the video recognition zone, FineCLIPER \cite{chen2024finecliper} using regenerated video captions for facial activities also improves the quality of the model. 
In the BFE problem, Li \etal \cite{obm} discovered that slightly larger building prompts can extract better footprints than entirely fitting box prompts. 
Another example is the application of the Large Language Model (LLM). 
RAG in LLM was another key tool to improve the final generated results\cite{RAG}. 
\textit{e.g.} Query2doc \cite{query2doc} use pseudo-documents prompting and concatenates them with the original query to improve predicting quality.

BFE problems solved by a promptable model like PolyFootNet must consider similar issues.
Researchers who proposed OBM, the former version of PolyFootNet, found that slightly inaccurate prompts can improve the performance of the model~\cite{obm}.
In this paper, PolyFootNet extended the "Prompting Test" to external models and studied three methods to extract footprints.

Except Fig.\ref{fig.multiSource} and Tab.\ref{tab:extr}, 
more interesting methods must help the model reach and exceed the ceiling performance of using Ground Truth prompts. 


\begin{figure}[!h]
\centering
\includegraphics[width=0.9\linewidth]{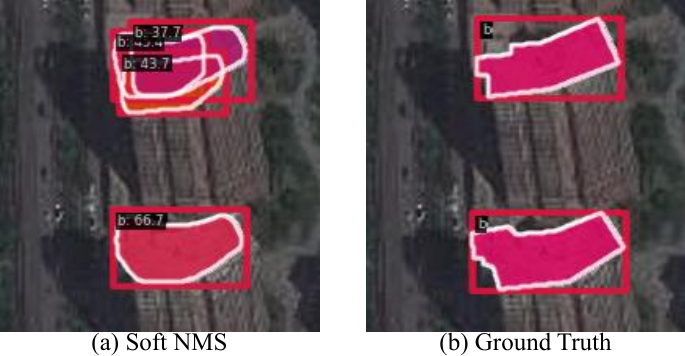}
\caption{Some mistake samples of roof extraction made by HTC and Ground Truth label}
\label{fig.roof_wrong}
\end{figure}

\subsection{The benefit of extracting footprints with \textit{building segmentation and offset}}
Considering building-related information was one of the contributions of this paper. 
For OBM and PolyFootNet, using building segmentation and offset robustly improves the quality of extracted footprints. 
We carefully analyzed the results of the predicted roof and building segmentation to find an explanation. 
We found that the edge of a roof and building facade is not that obvious compared with the edge between the building and background in one remote sensing image. 

\begin{figure}[!h]
\centering
\includegraphics[width=0.9\linewidth]{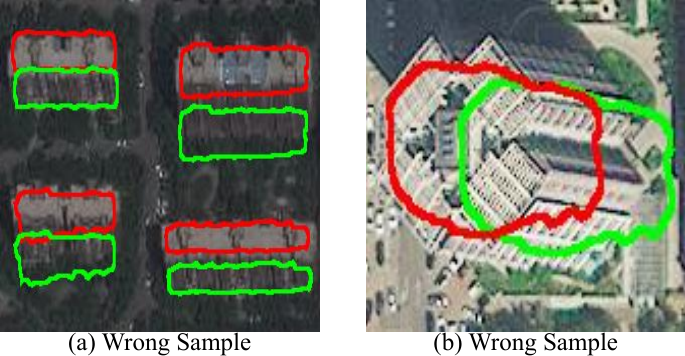}
\caption{Some mistake samples of roof extraction}
\label{fig.buhao}
\end{figure}
Fig.\ref{fig.buhao} shows some typical samples that predict false roofs due to the "edge problem". 
Low roof quality but relatively correct offsets finally lead to a poor quality footprint. 
The situation mentioned above can be improved by using building segmentation as in Fig.\ref{fig.intro}(c).


\subsection{Future work}
The contributions of PolyFootNet and the exploration of the multi-solution nature of the BFE problem extend beyond the scope of current experiments. In the future, designing more suitable visual prompts for PolyFootNet may be even more critical than developing new architectures, as effective prompting could significantly enhance model adaptability and performance.
Furthermore, the concept of multi-solution reasoning can be applied in reverse to address limitations in earlier datasets. \eg, the BANDON~\cite{pang2023detecting} dataset only provides roof and facade masks. By leveraging the multi-solution property, it becomes feasible to infer additional annotation types, such as footprints or building projections. This approach can support the construction of larger-scale off-nadir datasets, thereby facilitating the training of more robust and generalizable models for building footprint extraction.

%% file: sec/6_conclusion.tex
\input{chapters/Conclusion}

%% file: chapters/Conclusion.tex
\section{Conclusion}
\label{sec:Conclusion}
This paper presents PolyFootNet, a novel model for the BFE problem. 
Functionally, PolyFootNet can extract polygonal building footprints for off-nadir remote sensing images. 
Meanwhile, PolyFootNet effectively balances the prediction discrepancies between long and short offsets by introducing the SOFA. In addition, it explores the multi-solution nature of the BFE problem at the model level through combinatorial reasoning of building-related features.
The experimental results on three datasets demonstrate the superiority of our method.

%% file: author/authors.tex
\begin{IEEEbiography}[{\includegraphics[width=1in,height=1.25in,clip,keepaspectratio]{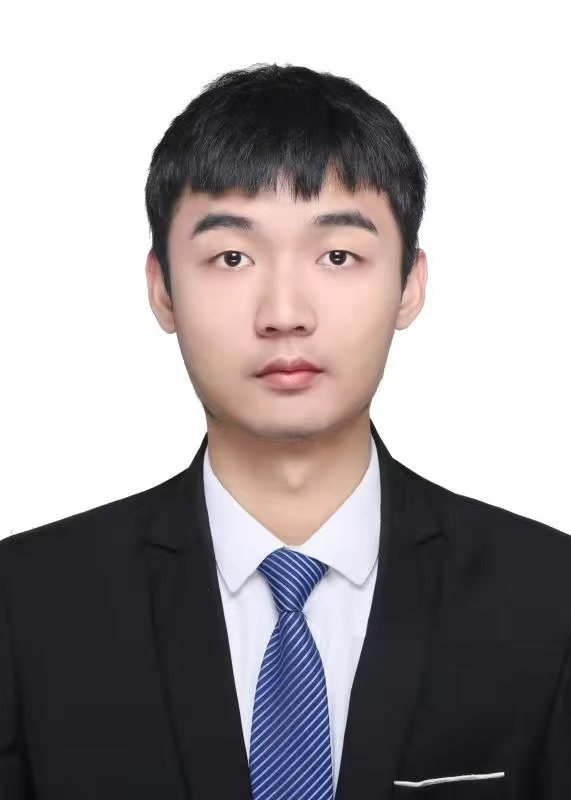}}]{Kai Li}\\
(Graduate Student Member, IEEE)
    received a bachelor’s degree in engineering, spatial information and digital technology from UESTC, Chengdu, China, in 2021. He is currently pursuing a PhD degree with UCAS, Beijing, China, supervised by \href{http://www.aircas.cas.cn/ykjs/lrld/201909/t20190903_5375259.html}{Zhongming Zhao} and \href{https://people.ucas.ac.cn/~0010249}{Yu Meng}.
    He also joined CityU of Hong Kong as a joint PhD student, and his supervisor is \href{https://zhaoxyai.github.io}{Xiangyu Zhao}. His research interests include remote sensing, computer vision, and machine learning.
\end{IEEEbiography}

\begin{IEEEbiography}[{\includegraphics[width=1in,height=1.25in,clip,keepaspectratio]{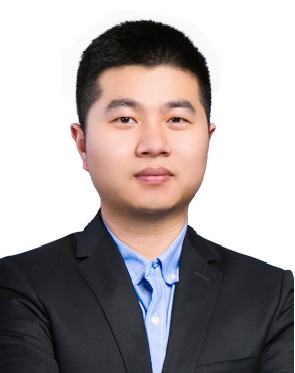}}]{Yupeng Deng}\\
    received the Ph.D. degree from the Aerospace Information Research Institute, Chinese Academy of Sciences, Beijing, China, in 2023.Now, he is a Post-Doctoral Researcher Supervised by Jianhua Gong. He is specialized in remote sensing and change detection.
\end{IEEEbiography}

\begin{IEEEbiography}[{\includegraphics[width=1in,height=1.25in,clip,keepaspectratio]{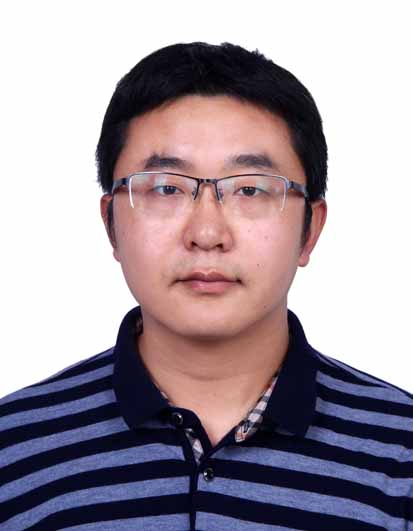}}]{Jingbo Chen}\\
(Member, IEEE) received the Ph.D.degree in cartography and geographic information systems from the Institute of Remote Sensing Applications, Chinese Academy of Sciences,Beijing, China, in 2011.\\
He is currently an Associate Professor with the Aerospace Information Research Institute, Chinese Academy of Sciences. His research interests cover intelligent remote sensing analysis, integrated application of communication, navigation, and remote sensing.
\end{IEEEbiography}

\begin{IEEEbiography}[{\includegraphics[width=1in,height=1.25in,clip,keepaspectratio]{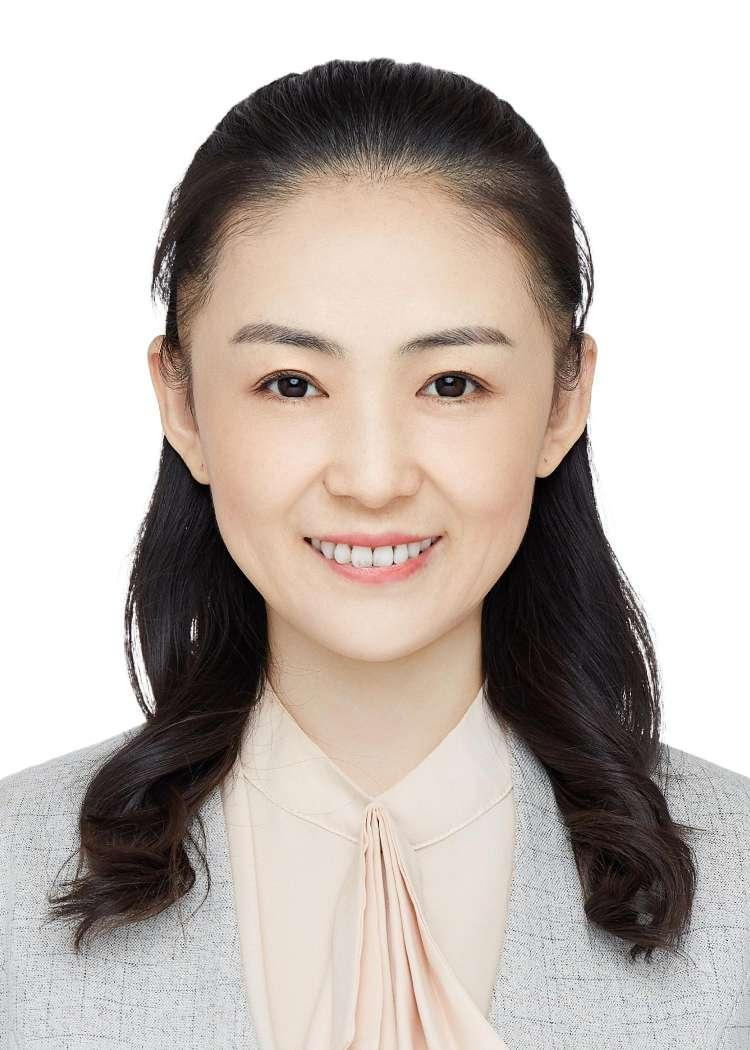}}]{Yu Meng}\\
     received the Ph.D. degree in signal and information processing from the Institute of Remote Sensing Applications, Chinese Academy of Sci-ences, Beijing, China, in 2008.\\
      She is currently a professor at the Aerospace Information Research Institute, Chinese Academy of Sciences. Her research interests include intelligent interpretation of remote sensing images,remote sensing time-series signal processing, and big spatial-temporal data application.\\
    Dr Meng serves as an editor and board member of the National Remote Sensing Bulletin, Journal of Image and Graphics.

\end{IEEEbiography}

\begin{IEEEbiography}[{\includegraphics[width=1in,height=1.25in,clip,keepaspectratio]{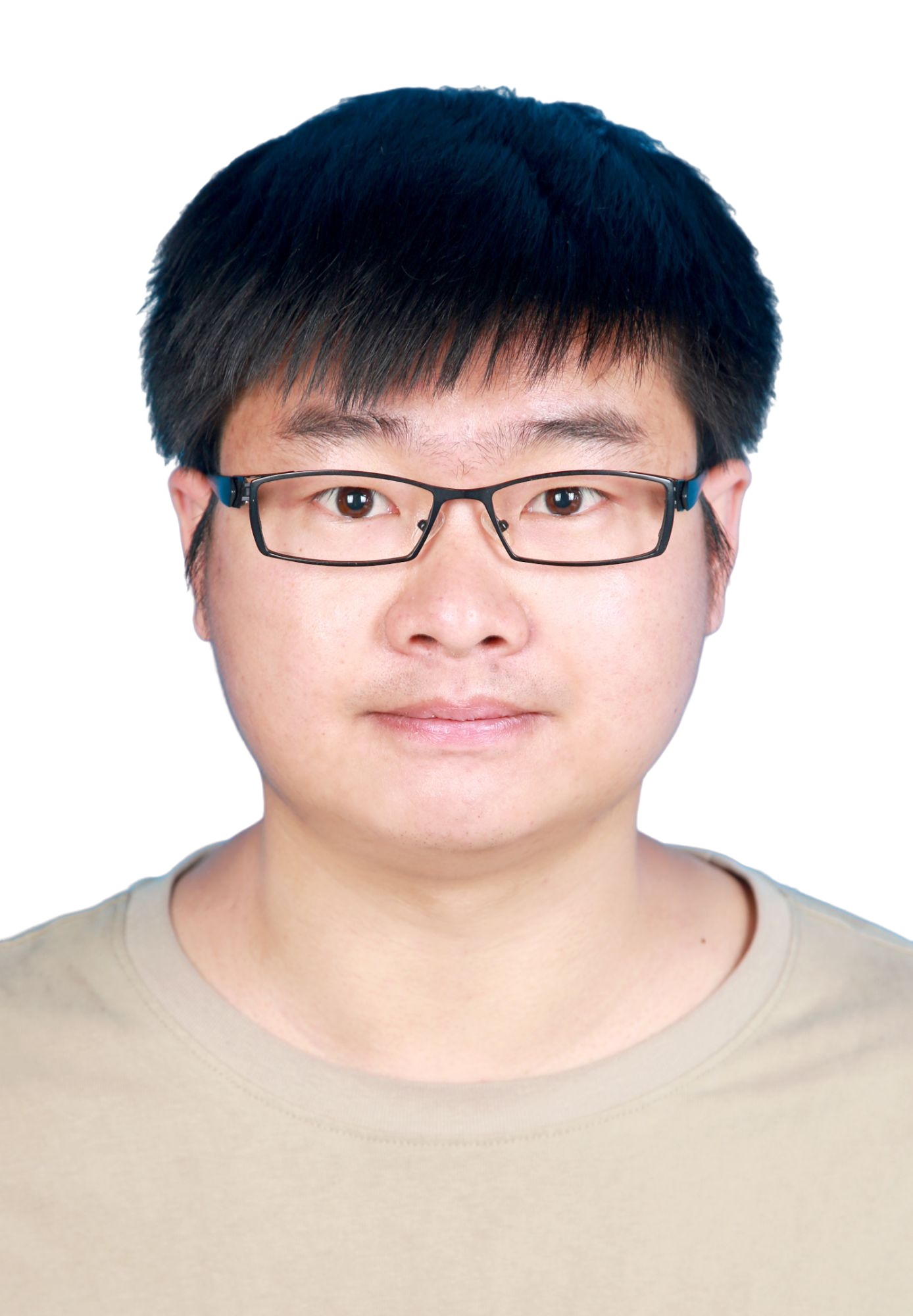}}]{Zhihao Xi}\\
    received the B.S. degree from the Wuhan University of Technology, Wuhan, China, in 2019, and the Ph.D. degree from the Aerospace Information Research Institute, Chinese Academy of
Sciences (CAS), Beijing, China, in 2024.
   He is currently an Assistant Professor with the Aerospace Information Research Institute, CAS. His research interests include computer vision, domain adaptation, and remote sensing image interpretation.
\end{IEEEbiography}

\begin{IEEEbiography}[{\includegraphics[width=1in,height=1.25in,clip,keepaspectratio]{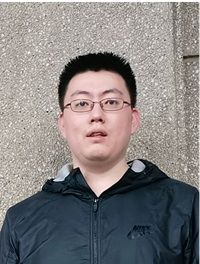}}]{Junxian Ma}\\
    received the bachelor’s degree from Peking University, Beijing, China, in 2020. He is currently pursuing the Ph.D. degree with the Uni-versity of Chinese Academy of Sciences, Beijing.He focuses on image generation and conditional video generation.
\end{IEEEbiography}

\begin{IEEEbiography}[{\includegraphics[width=1in,height=1.25in,clip,keepaspectratio]{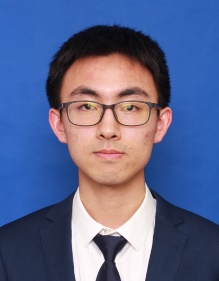}}]{Chenhao Wang}\\
   received the bachelor’s degree from the University of Electronic Science and Technology of China, Chengdu, China, in 2022. He is currently pursuing the Ph.D. degree with the University of Chinese Academy of Sciences, Beijing, China.His research focuses on building extraction from remote sensing images.
\end{IEEEbiography}

\begin{IEEEbiography}[{\includegraphics[width=1in,height=1.15in,clip,keepaspectratio]{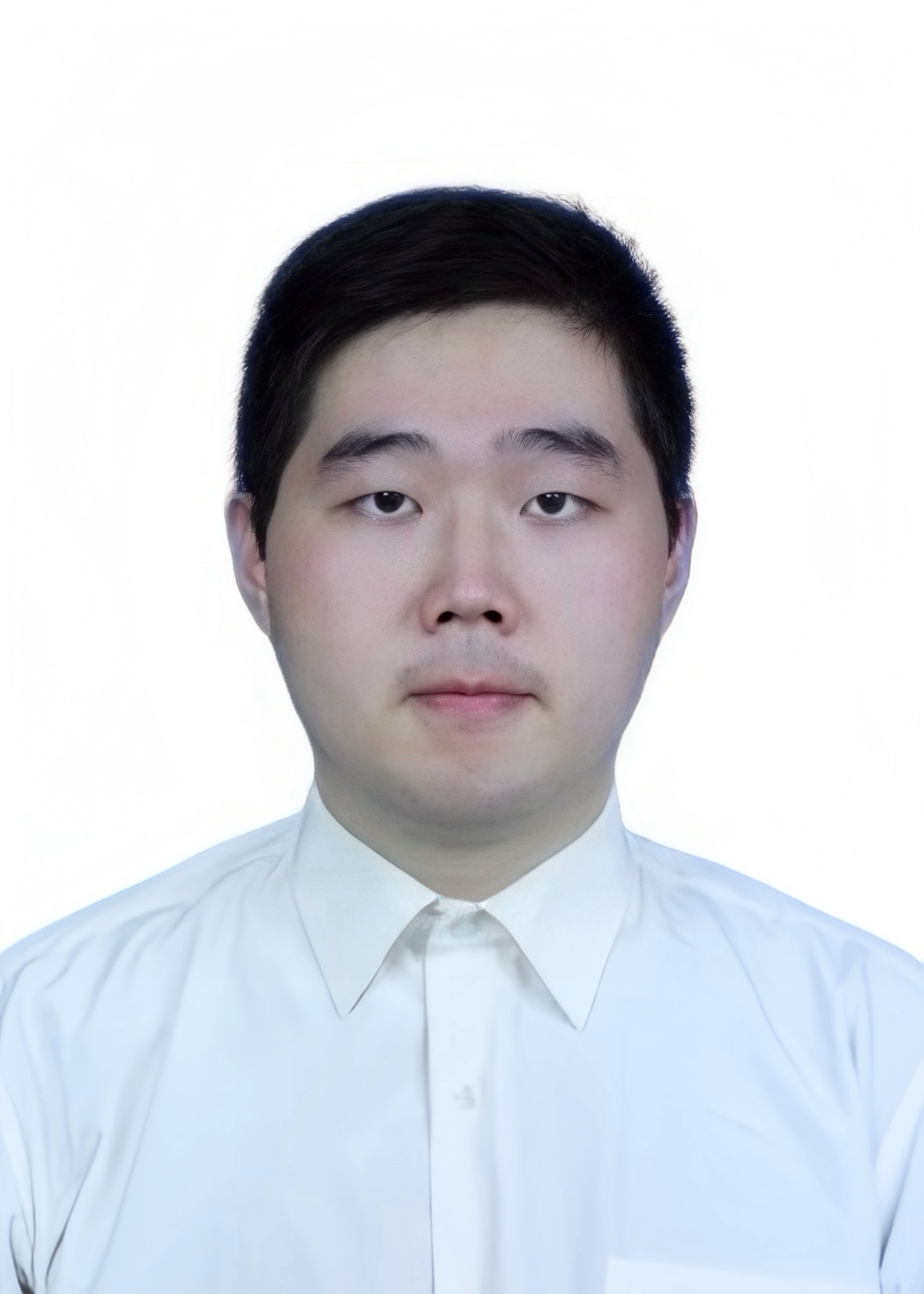}}]{Maolin Wang}
Maolin Wang received an M.Phil degree from the School of Computer Science and Engineering, University of Electronic Science and Technology of China, Sichuan, China, 2021. He is currently pursuing a Ph.D. degree in data science at the City University of Hong Kong, HKSAR, China. His research interests include tensor, graph neural networks, and their applications.
\end{IEEEbiography}

\begin{IEEEbiography}[{\includegraphics[width=1in,height=1.25in,clip,keepaspectratio]{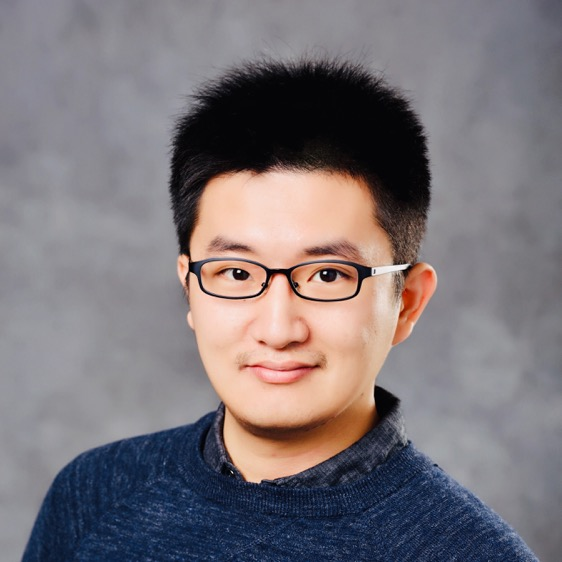}}]{Xiangyu Zhao}\\
is an assistant professor at the
school of Data Science at the City University of Hong
Kong (CityU). Before CityU, he completed his
Ph.D. at Michigan State University. His current re-
search interests include data mining and machine
learning, especially on Reinforcement Learning
and its applications in Information Retrieval. He
has published papers in top conferences (e.g.,
KDD, WWW, AAAI, SIGIR, ICDE, CIKM, ICDM,
WSDM, RecSys, ICLR) and journals (e.g., TOIS,
SIGKDD, SIGWeb, EPL, APS). His research received ICDM’21 Best-ranked Papers, Global Top 100 Chinese New Stars in AI, CCF-Tencent Open Fund, Criteo Research Award, and Bytedance Research Award. He serves as top data science conference (senior) program committee members and session chairs (e.g., KDD,AAAI, IJCAI, ICML, ICLR, CIKM), and journal reviewers (e.g., TKDE,TKDD, TOIS, CSUR). He is the organizer of DRL4KDD@KDD’19,DRL4IR@SIGIR’20, 2nd DRL4KD@WWW’21, 2nd DRL4IR@SIGIR’21,and a lead tutor at WWW’21 and IJCAI’21. More information about him can be found at https://zhaoxyai.github.io/.
\end{IEEEbiography}